\documentclass[sigconf]{acmart}

\preto{\abstractkeywords}{\nolinenumbers}

\usepackage{enumerate}
\usepackage{multirow} 
\usepackage{booktabs}
\usepackage{siunitx}
\usepackage{array} 
\usepackage{caption}
\usepackage{graphicx}
\usepackage{amsmath}
\AtBeginDocument{%
	}

\renewcommand\footnotetextcopyrightpermission[1]{}
\author{Yutong Chen}
\affiliation{%
	\institution{University of Science and Technology of China}
	\city{Hefei}
	\state{Anhui}
	\country{China}
}
\email{yutongchen@mail.ustc.edu.cn}
\author{Zhang Wen}
\affiliation{%
	\institution{Ocean University of China}
	\city{Qingdao}
	\state{Shandong}
	\country{China}
}
\email{wenzhang@stu.ouc.edu.cn}
\author{Chao Wang}
\affiliation{%
	\institution{University of Science and Technology of China}
	\city{Hefei}
	\state{Anhui}
	\country{China}
}
\email{cswang@ustc.edu.cn}

\author{Lei Gong}
\affiliation{%
	\institution{University of Science and Technology of China}
	\city{Hefei}
	\state{Anhui}
	\country{China}
}

\email{leigong0203@ustc.edu.cn}

\author{Zhongchao Yi}
\affiliation{%
	\institution{University of Science and Technology of China}
	\city{Hefei}
	\state{Anhui}
	\country{China}
}
\email{zhongchaoyi@mail.ustc.edu.cn}

\begin{document}
	
	\title{PriorNet: A Novel Lightweight Network with Multidimensional Interactive Attention for Efficient Image Dehazing}
	


	\begin{abstract}
		Hazy images degrade visual quality, and dehazing is a crucial prerequisite for subsequent processing tasks. Most current dehazing methods rely on neural networks and face challenges such as high computational parameter pressure and weak generalization capabilities. This paper introduces PriorNet—a novel, lightweight, and highly applicable dehazing network designed to significantly improve the clarity and visual quality of hazy images while avoiding excessive detail extraction issues. The core of PriorNet is the original Multi-Dimensional Interactive Attention (MIA) mechanism, which effectively captures a wide range of haze characteristics, substantially reducing the computational load and generalization difficulties associated with complex systems. By utilizing a uniform convolutional kernel size and incorporating skip connections, we have streamlined the feature extraction process. Simplifying the number of layers and architecture not only enhances dehazing efficiency but also facilitates easier deployment on edge devices. Extensive testing across multiple datasets has demonstrated PriorNet's exceptional performance in dehazing and clarity restoration, maintaining image detail and color fidelity in single-image dehazing tasks. Notably, with a model size of just 18Kb, PriorNet showcases superior dehazing generalization capabilities compared to other methods. Our research makes a significant contribution to advancing image dehazing technology, providing new perspectives and tools for the field and related domains, particularly emphasizing the importance of improving universality and deployability.
	\end{abstract}

	
	\maketitle
	
	\section{Introduction}
	
	Images captured in foggy or hazy conditions often exhibit reduced visual quality, characterized by lower contrast and color distortion \cite{4587643}. This degradation not only negatively impacts the visual experience, but also poses significant challenges for computer vision applications, such as image classification and object detection, which largely rely on datasets composed of clear images \cite{ren2016single}, \cite{blasinski2018optimizing}. As a result, images obtained in hazy conditions can adversely affect the performance of advanced visual tasks. To ensure these tasks are completed effectively, it is imperative that input images are either free of haze or minimally impacted by it. In recent years, the field of image dehazing, which focuses on restoring clear scenes from hazy images, has received widespread attention from both the academic and industrial sectors.
	
	Initial research in dehazing primarily relied on atmospheric scattering models, which offer a simplified approximation of fog effects as shown in the equation \cite{mccartney1976optics}, \cite{meng2013efficient}, \cite{narasimhan2000chromatic}:
	\begin{equation}
		\label{AMS}
		I(z) = J(z)t(z) + A(1 - t(z))
	\end{equation}
	In this model,I(z) represents the observed hazy image, A is the global atmospheric light, t(z) is the medium transmission map, and J(z) denotes the haze-free image. Early studies highlighted the practical applicability of this model, with the Dark Channel Prior (DCP) method \cite{he2010single} emerging as a notably effective solution in real-world conditions due to its computational efficiency. However, methods based on priors may encounter challenges due to inaccurate estimations of the transmission map, often resulting from violations of prior assumptions in practical scenarios, potentially limiting their effectiveness. This can lead to color distortion or the presence of halos and light patches around edges, impacting the progress of subsequent tasks.

	Recent advancements in dehazing research have unequivocally affirmed the superior efficacy of deep neural networks in image dehazing techniques, encompassing both CNN \cite{cai2016dehazenet}, \cite{li2017aod}, \cite{li2020deep}, \cite{zhang2018densely} and Transformer \cite{qiu2023mb}, \cite{song2023vision}, \cite{liu2023visual} architectures. With the increasing depth of these networks and the integration of various attention mechanisms throughout the dehazing process, there has been a notable enhancement in network performance \cite{lu2023mixdehazenet}, \cite{chen2024dea}. However, this enhancement comes at the cost of escalated algorithmic complexity. This complexity introduces two primary challenges: an exponential increase in the parameter count, elevating the costs associated with training and inference, and a decline in the generalization capability. While these sophisticated algorithms excel in specific datasets, their robustness diminishes when applied to alternative datasets or under the fluctuating conditions of the real world. This situation highlights the pressing need to explore strategies that maintain computational efficiency while improving the reliability and broad applicability of dehazing algorithms.
	
	To address the challenges in the field of dehazing, we introduce a novel, lightweight, and highly generalizing dehazing network (named PriorNet). This approach harnesses neural network techniques to extract fundamental features from actual images, followed by the application of the atmospheric scattering model for image restoration. In the extraction phase, the method leverages the nonlinear abstraction capabilities of neural networks to extract haze characteristics that are challenging to quantify or derive. For restoration, it emphasizes the application of a physical model,namely AMS \cite{narasimhan2003contrast}, to achieve accurate dehazing. DCP highlights the extremely low intensity in at least one RGB channel of certain pixels, while also observing that haze distribution within images exhibits regional characteristics - neither uniformly spread across the entire image nor presenting pixel-level variances. Thus, there arises an urgent need for a new type of attention mechanism that is neither as coarse as channel attention nor as overly detailed as spatial attention. In response, we propose the Multidimensional Interactive Attention (MIA), designed to capture the broad haze features across different regions, avoiding the high computational strain and reduced generalizability associated with complex attention mechanisms, thereby achieving simplicity and efficiency. Moreover, by utilizing a shallow network architecture and uniform convolutional kernels, our network focuses on extracting the most crucial information, preventing excessive detail extraction for more effective dehazing outcomes.Furthermore, in the feature extraction phase, our network deviates from the traditional approach of utilizing various convolutional kernel sizes. Instead, it opts for a single kernel size, coupled with the use of skip connections, to capture a more comprehensive array of physical information. By constraining our network to 5 convolutional layers and 3 concatenation layers, we not only ensure the thoroughness of information capture at a consistent scale, avoiding the extraction of excessive detail, but also significantly reduce the total parameter count of the network. This streamlined architecture facilitates easier deployment on edge devices, optimizing both efficiency and practicality.During the restoration phase, operations are carried out on a per-pixel basis, details of which will not be elaborated further here.

	Our contributions in advancing image dehazing technology focus on three primary innovations:
	\begin{itemize}
		\item \textbf{Innovative Network Design (PriorNet):} A novel, lightweight dehazing network distinguished by its remarkable generalization capabilities. PriorNet is engineered to efficiently process hazy images, enhancing clarity and visual quality without the need for excessive detail extraction.
		\item \textbf{Multidimensional Interactive Attention (MIA):
		} An innovative attention mechanism tailored to efficiently capture widespread haze features without the computational burden and generalization issues of complex systems, steering towards a more effective dehazing process.
		\item \textbf{Optimized Feature Extraction and Restoration:} By adopting a uniform convolutional kernel size and utilizing skip connections, we streamline the feature extraction process. Our network, with its reduced layer count and simplified architecture, not only enhances the dehazing efficiency but also supports easier deployment on edge devices.
	\end{itemize}

	\section{RELATED WORK}
	
	\subsection{Attention Mechanism}
	In deep learning, the attention mechanism has emerged as a pivotal technique \cite{bahdanau2014neural}, \cite{dosovitskiy2020image}, acclaimed for its adeptness in heightening the model's focus on pivotal elements within the input data \cite{liu2019griddehazenet}, \cite{wang2019aagan}, \cite{jiang2021haze}. Self-attention mechanisms elevate model performance by pinpointing long-distance interactions among inputs, thereby capturing nuances often missed by conventional approaches \cite{vaswani2017attention}. Channel attention scrutinizes the dynamics of feature activations across channels \cite{hu2018squeeze}, spotlighting the relational importance of different features, whereas spatial attention assesses the criticality of information's spatial layout \cite{zhu2019empirical}, optimizing the model's perceptual field. The advent of numerous derivatives of these core mechanisms has further broadened the spectrum of attention module designs, significantly enhancing model interpretability and adaptability in processing complex datasets. 
	
	Notwithstanding the apparent benefits of attention mechanisms, they unavoidably introduce a significant number of parameters and computational overheads. Furthermore, in the endeavor to highlight essential information, there is a potential for models to capture an overabundance of redundant information, potentially leading to overfitting or a decline in performance in real-world applications.
	
	To tackle these challenges, we draw inspiration from the domains of physical models and mathematical statistics, merging channel and spatial attention mechanisms to innovatively propose the Multidimensional Interactive Attention (MIA). MIA unites the strengths of traditional channel and spatial attention mechanisms, effectively extracting fundamental haze features across different image regions from the input data. With its simplified structure, MIA not only alleviates computational load and training costs but also avoids the excessive extraction of information, focusing instead on capturing key details. 
	
	\subsection{Single Image Dehazing}
	
	Initial studies in image dehazing were predominantly based on mathematical statistics and physical methodologies, frequently utilizing atmospheric scattering models or leveraging specific handcrafted priors \cite{berman2016non}, \cite{chen2016robust}. Although these methods are grounded in solid theoretical principles, their real-world applicability often encounters limitations due to the precision of assumed prior conditions, resulting in less-than-ideal dehazing outcomes in complex environmental situations. The gap between the theoretical environmental assumptions and the realities of actual conditions can markedly affect the effectiveness of dehazing techniques.
	
	With the advancement of deep learning technologies, current research \cite{qin2020ffa}, \cite{lu2023mixdehazenet}, \cite{wang2024image} has primarily shifted towards data-driven approaches, especially those employing Convolutional Neural Networks (CNNs) or Transformer architectures to estimate atmospheric lighting and transmission maps. The incorporation of numerous attention mechanisms has significantly enhanced model performance. These methods construct dehazing models by learning from vast datasets, effectively improving the accuracy of dehazing. However, an increase in model complexity also means a rise in the number of parameters and computational load, which often affects the generalization ability across different datasets and in real-world application scenarios.
	
	In response to the limitations inherent in existing dehazing methods, we introduce an innovative dehazing network design. Grounded in the atmospheric scattering model, our approach diverges from traditional CNN methodologies in feature extraction. Our model incorporates a novel attention mechanism and abstracts real-world conditions through a shallow network architecture, employing single-size convolutional kernels and skip connections, with a heightened focus on the application of physical models and mathematical principles. Consequently, our network significantly enhances adaptability and resilience to changes in real-world environments, while dramatically reducing the number of parameters and computational demands.
	
	\begin{figure*}
		\centering
		\includegraphics[width=\textwidth]{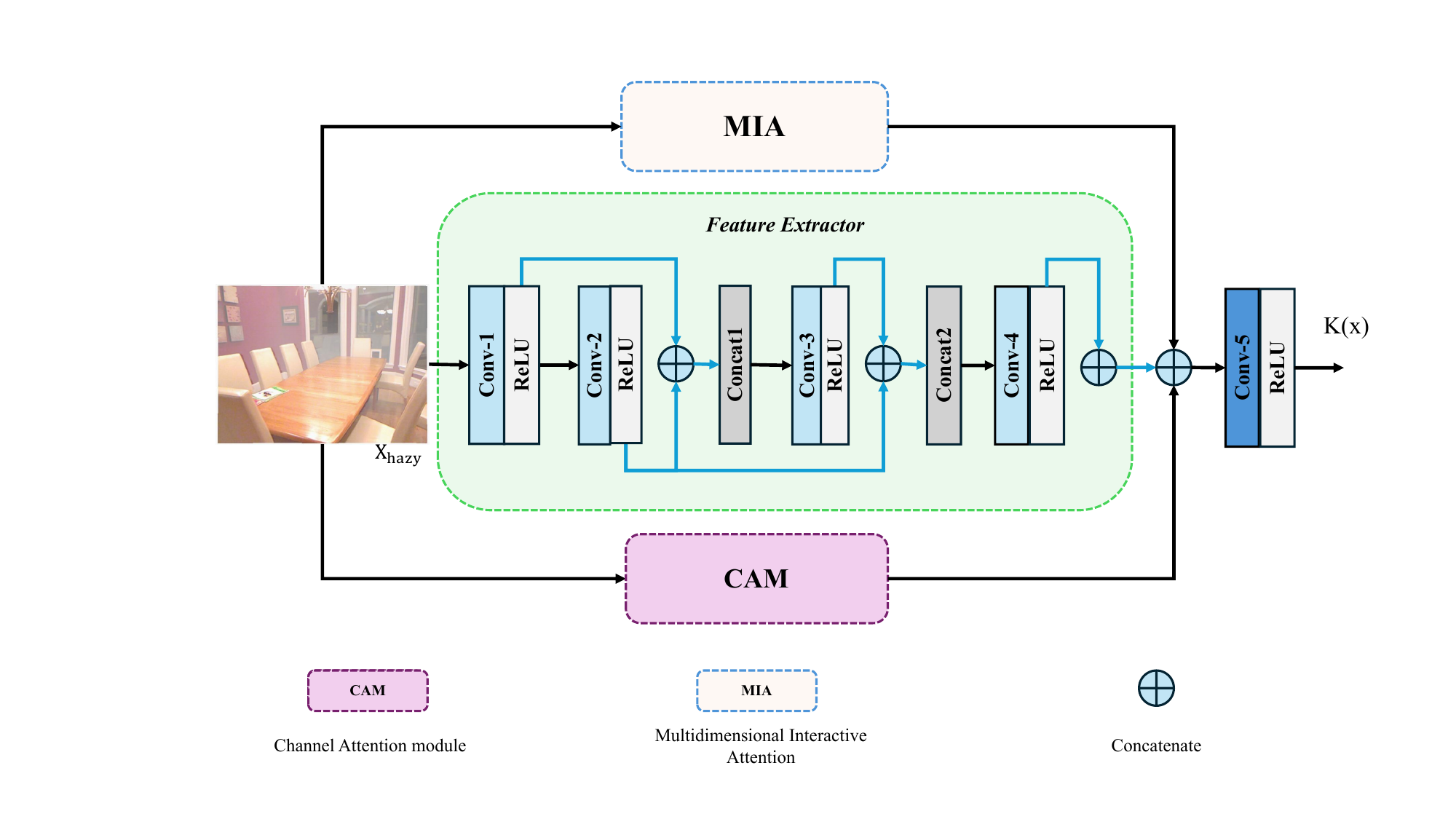}
		\caption[sacle = 0.05]{Architecture of PriorNet}
		\label{fig:architechture_priornet}
	\end{figure*}
	
	\section{Methods}
	
	\subsection{Physics Model}
	In environments impacted by haze, image quality is significantly compromised, presenting challenges at both global and local levels. Assuming a uniform distribution of haze within the atmosphere allows for an initial understanding of the overall haze density within images. However, the degradation of image quality is not solely influenced by this global haze density but is also affected by variations in lighting conditions and the distance between objects and the camera. This necessitates the segmentation of the dehazing problem into global and local components for a more effective resolution.
	
	The vast majority of dehazing research, ranging from early statistical methods to contemporary deep neural network algorithms, is grounded on the atmospheric scattering model. Building on this model, namely equation \ref{AMS}, the image restoration process can be reformulated as \cite{li2017aod}:
	\begin{equation}
		\label{AMS_Dehaze}
		J(x) = K(x) \cdot I(x) - K(x) + b
	\end{equation}
	where
	\begin{equation}
		\
		K(x) = \frac{1}{t(x)} \cdot \left(\frac{I(x) - A}{I(x) - 1}\right) + \frac{(A - b)}{I(x) - 1}
	\end{equation}
	
	Here, \(K(x)\) acts as a scaling factor adjusting the intensity of the observed image pixels, and \(b\) is a bias term introduced to mitigate potential errors. The scaling factor \(K(x)\) accounts for the attenuation of light caused by the haze, and the bias \(b\) addresses any ambient light potentially skewing the intensity levels.
	
	In traditional image dehazing methods, the restoration process heavily relies on mathematical and statistical approaches to estimate atmospheric light \(A\) and transmission \(t(x)\) separately before proceeding to image recovery. For instance, the Dark Channel Prior (DCP) algorithm initially calculates the dark channel image, uses it to estimate atmospheric light \(A\), and then combines the dark channel image with the estimated atmospheric light to deduce the transmission \(t(x)\). Finally, it utilizes the atmospheric scattering model (AMS) to recover the haze-free image. In contrast, modern methods based on deep learning adopt a more straightforward approach by directly extracting the scaling factor \(K(x)\), instead of separately determining the transmission map \(t(x)\), atmospheric light \(A\), and bias \(b\). This approach not only simplifies the processing steps but also leverages the neural network's strong capability in fitting nonlinear problems, effectively overcoming the limitations in flexibility and adaptability of traditional methods. Thus, it allows for a more accurate restoration of the original haze-free image \(J(x)\).
	
	\subsection{Network Design}
	Building upon our previous discussion, we segregate the dehazing challenge into two principal categories: overarching global concerns and nuanced local disparities. To adeptly address these divisions, we present PriorNet, a innovative and ultra-lightweight network framework specifically engineered to tackle the complexities of the dehazing process. Perfectly aligning with the forefront of advancements in deep learning for dehazing, PriorNet encompasses two essential components: a sophisticated deep learning estimation module for \(K(x)\) analysis and an image restoration module that employs fundamental operations for the reconstruction of dehazed images. This restoration module relies heavily on a methodical direct estimation of \(K(x)\) for reconstructing images on a pixel-by-pixel basis, a methodology thoroughly vetted in prior studies. Therefore, we opt not to reiterate on this process. The essence of our network’s design resides in the \(K(x)\) estimation module, which cleverly merges a convolutional extraction pathway with a multidimensional interactive attention (MIA) framework, as depicted in Figure \ref{fig:architechture_priornet}. This estimation module inputs hazy images to directly output feature maps. In this section, we aim to meticulously delve into the specifics of these two foundational elements.
	
	\subsection{Multidimensional Interactive Attention} In response to the inherent properties of haze, we have endeavored to integrate modules adept at gleaning both overarching and nuanced details. This initiative led to the development of the Multidimensional Interactive Attention (MIA). MIA leverages a traditional channel attention mechanism to discern global fog characteristics, while it harnesses a uniquely devised spatial attention mechanism to accurately extract information from pivotal regions within the image.
	
	The channel attention mechanism operates according to the principle outlined in Equation \ref{channel_attention}. This mechanism employs a combination of average pooling and max pooling to process the feature map \(F\), subsequently integrating the outcomes through a multi-layer perceptron (MLP) with shared weights. The MLP consists of two weight matrices, \(W_0\) and \(W_1\), and incorporates a ReLU activation function post the first weight layer. The final channel attention map, \(M_c(F)\), is generated by applying the sigmoid function \(\sigma\) to the aggregated MLP outputs, offering a refined method to dynamically emphasize informative features.
	
	\begin{equation}
		\label{channel_attention}
		M_c(F) = \sigma\left(W_1\left(W_0\left(\text{AvgPool}(F)\right) + W_0\left(\text{MaxPool}(F)\right)\right)\right)
	\end{equation}
	
	Here, \(\sigma\) represents the sigmoid function, indicating a non-linear transformation applied to the MLP outputs. The matrix \(W_0 \in \mathbb{R}^{r \times c}\) acts as the initial transformation layer within the MLP, followed by \(W_1 \in \mathbb{R}^{c \times c / r}\), which further refines the attention mechanism's output.

	\begin{figure*}
		\centering
		\includegraphics[scale=0.5]{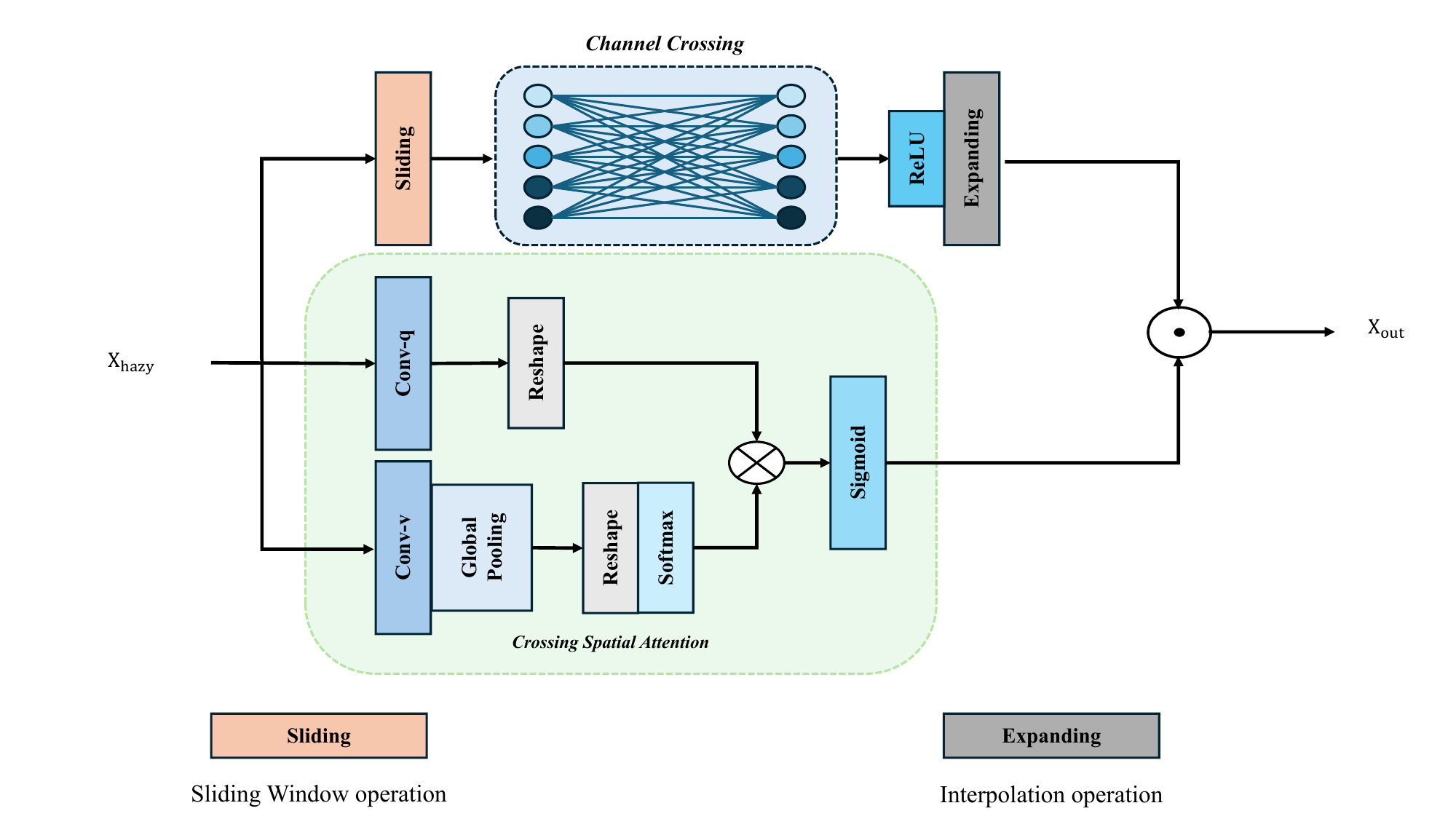}
		\caption{Architecture of the MIA }
		\label{fig:sophisticated_spatial_attention}
	\end{figure*}
	We have engineered an advanced spatial attention mechanism specifically designed to identify local features, structured in three primary sections, as illustrated in Figure \ref{fig:sophisticated_spatial_attention}.
	
	The initial phase is the local spatial attention module, which employs a sliding window technique to partition the hazy image, thereby isolating significant local features. A fully connected layer is then employed to bolster the interchange of channel information among these features. Once this process is complete, the feature vectors are expanded to their original dimensions.
	\begin{equation}
		\label{equation_local}
		F_{local}\left( x_{hazy} \right) =F_{ex}\cdot \left( F_{ReLU}\cdot\left( F_{MLP}\left( F_{sw}\left( x_{hazy} \right) \right) \right) \right)
	\end{equation}
	
	In Equation \ref{equation_local}, The expansion operator, denoted as $F_{ex}$, ensures that features are resized back to their original dimensions. $F_{ReLU}$ signifies the ReLU activation function, $F_{MLP}$ refers to a fully connected layer, and $F_{sw}$ is the sliding window average pool operator, characterized by a stride of 4 and a window size of 8x8.
	
	Subsequently, our spatial cross-attention framework is divided into two parts: initially, feature enhancement is achieved through a fully connected layer; following this, channel and spatial features are derived using average pooling and convolution operations. These features are then combined through a reshaping and cross-multiplication strategy, effectively merging spatial and channel data. The local spatial features and these cross features are then integrated through a multiplicative fusion.
	
	\begin{equation}
		F_{SP}\left( x \right) =F_{SG}\left[ \sigma _3\left( F_{SM}\left( \sigma _1\left( F_{GP}\left( W_q\left( x \right) \right) \right) \right) *\sigma _2\left( W_2\left( x \right) \right) \right) \right]
	\end{equation}

	In this context, \(F_{sp}\) $\in$ \(\mathbb{R}^{1\times H \times W}\), with \(F_{GP}\) signifying the global pooling layer, as delineated in Equation \ref{equation5}.
	\begin{equation}
		\label{equation5}
		F_{GP}\left( x \right) =\frac{1}{H*W}\sum_{i=1}^H{\sum_{j=1}^W{x}\left( :,i,j \right)}
	\end{equation}
	The operation of the softmax function is represented by \(F_{SM}(\cdot)\), detailed in Equation \ref{equation6}, while the sigmoid function is encapsulated by \(F_{SG}(X_{pro}(\cdot))\), as expounded in Equation \ref{equation7}.

	\begin{equation}
		\label{equation6}
		F_{SM}\left( x_{pro}\left( x_i,y_i,c_i \right) \right) =\frac{e^{x_{pro}\left( x_i,y_i,c_i \right)}}{\sum_{x_j\epsilon \left( 0,H \right)}{\sum_{y_j\epsilon \left( 0,H \right)}{\sum_{z_j\epsilon \left( 0,H \right)}{e^{x_{pro}\left( x_j,y_j,c_j \right)}}}}}
	\end{equation}
	
	\begin{equation}
		\label{equation7}
		F_{SG}\left( x_{pro}\left( x_i,y_i,c_i \right) \right) =\frac{1}{1+e^{-x_{pro}\left( x_i,y_i,c_i \right)}}
	\end{equation} 
	The '$\times$' symbolizes the matrix product operator, while $W_q$ and $W_v$ are tensor reshape operators. $\sigma1$, $\sigma2$, $\sigma3$ are additional tensor reshape operators.
	
	The finale of this process involves the amalgamation of these two feature sets. The local features serve to enhance the cross features, shown as equation \ref{equation4}.
	
	\begin{equation}
		\label{equation4}
		F_{atten}\left( x_{hazy} \right) =F_{SG}\left( F_{cocal}\left( x_{hazy} \right) \odot F_{sp}\left( x_{hazy} \right) \right)  
	\end{equation}
	
	Ultimately, all the extracted and combined features are brought together through a convolutional procedure, resulting in the production of the ultimate, polished image.
	
	\begin{figure*}
		\centering
		\includegraphics[scale=0.5]{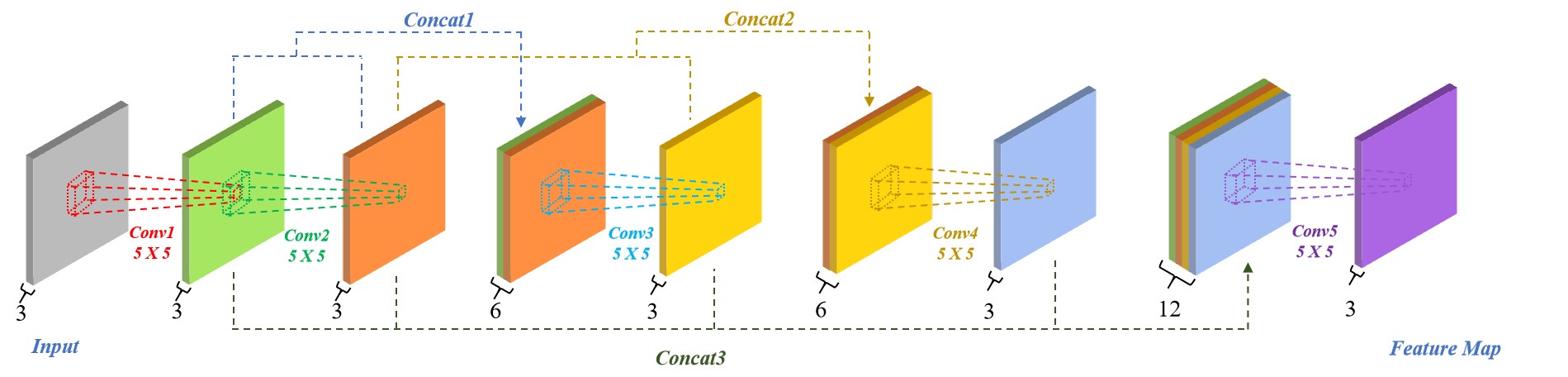}
		\caption{Architecture of Convolutional Feature Extraction Module}
		\label{fig:cnn}
	\end{figure*}

	\subsection{Convolutional Feature Extraction Module}
	Leveraging the structural insights of AOD-Net \cite{li2017aod} and the profound analytical rigor of statistical methods exemplified by DCP \cite{he2010single}, our approach is grounded in the robust principles of mathematics and physics, aiming to demystify the complexities inherent in haze characterization. Consequently, we've streamlined our convolutional feature extraction architecture, as illustrated in Figure \ref{fig:cnn}, which consists of five convolutional layers paired with three Concat operations, ensuring a thorough extraction of features. Diverging from traditional pixel-centric methods, we posit that haze predominantly exhibits a regional distribution. This pivotal realization steered us towards standardizing our convolutional kernel size, thereby sidestepping the need for a diverse arsenal of convolutional operations. After meticulous evaluation, we adopted a 5x5 kernel size, identified as the ideal choice for enhancing network performance due to its well-balanced receptive field.
	
	Furthermore, the "concat1" layer skillfully merges features from "conv1" and "conv2", while "concat2" and "concat3" seamlessly integrate features from subsequent convolutional layers, with each layer's contribution distinctly highlighted in different colors within the figure \ref{fig:cnn}for enhanced visual clarity. This multi-scale feature fusion methodology is key to mining pivotal information from across the entire image, substantially elevating the network’s efficiency in discerning and mitigating haze. These deliberate design decisions, including an optimized network architecture with a reduced count of layers and parameters and a strategic choice to limit convolutional kernel sizes, culminate in PriorNet. This platform embodies a sophisticated yet highly effective dehazing solution, deeply informed by an understanding of the regional nature of haze, setting a new standard in the field of image dehazing.
	
	\subsection{Loss Function}
	In our model, we utilize the Mean Squared Error (MSE) as the loss function to quantify the discrepancies between the original and the output images, defined as:
	\begin{multline}
		Loss_{mse} = \frac{1}{C*H*W} \sum_{x_i \in \left[ 0,H \right]} \sum_{y_i \in \left[ 0,W \right]} \sum_{c_i \in \left[ 0,C \right]} \\
		\left[ x_{gt}\left( x_i,y_i,c_i \right) - x_{pro}\left( x_i,y_i,c_i \right) \right]^2
	\end{multline}

	where \(X_{gt}\) represents the clean, haze-free image, \(X_{pro}\) is the network's output, \(H\) is the height of the image, \(W\) is the width, and \(C\) denotes the channels.
	
	Additionally, to enhance the human perceptual quality of the processed images, we integrate a perceptual loss function, utilizing a pre-trained Vgg16 network as the perceptual evaluator \cite{qu2020crack}:
	\begin{equation}
		Loss_{perception}=\frac{1}{N}\sum_{i=1}^N{\left( F_{Vgg16}\left( x_{gti} \right) -F_{Vgg16}\left( x_{pro\ i} \right) \right) ^2}
	\end{equation}
	with \(N\) being the batch size and \(F_{Vgg16}\) the Vgg16 network equipped with pre-trained weights.
	
	To align the processed images closely with their originals, we adopt a coefficient \(beta=0.1\) to temper the perceptual loss's effect:
	\begin{equation}
		Loss=Loss_{mse}+\beta \cdot Loss_{perception}
	\end{equation}
	This formulation not only aims at minimizing the direct errors between images but also at improving their realism and visual appeal to the human eye, while preserving image detail and quality.
	
	\section{EXPERIMENTS}
	
	In this section, we will firstly describe the experimental setup. After that, the results of the experiments and an ablation study of PriorNet will be illustrated.
	
	\subsection{Experiment Settings}
	
	\paragraph{Datasets.} To precisely assess the generalization capacity of various dehazing models, our research utilized two distinct datasets: "Haze4k" \cite{liu2021synthetic} and "Hazy\_NYU\_Depth-V2" \cite{li2017aod}. For ease of reference in further discussions, the latter is abbreviated as "Hazy\_NYU\\\_DepthV2", which consists of hazy images from the NYU Depth V2 dataset and has been previously utilized for testing within the AOD-Net framework. It is critical to mention that the designations adopted in this study are solely for the convenience of expression and do not correspond to their official titles. The choice of the "Haze4k" dataset is justified by its prevalent use in the training of dehazing algorithms, whereas the "Hazy\_NYU\_DepthV2", derived from the NYU Depth V2 dataset, facilitates a more extensive spectrum of experimental automation by supporting tasks in both dehazing and recognition domains.

	\paragraph{Implementation Details.} Our methodology employs a dual-source approach for testing models. Specifically, we utilize models in their original form if they were previously trained on the "Haze4K" dataset, adhering to the configurations established by their creators. For models that lack prior training on "Haze4K," we undertake a rigorous retraining process on this dataset to ensure they are well-adapted for evaluation. This enables us to establish the most effective parameters for these models, tailoring their performance to our assessment criteria. All models are evaluated against the "Hazy\_NYU\_DepthV2" test set to ensure a fair and accurate comparison of their real-world effectiveness. This strategy allows us to assess models from both sources under consistent conditions, offering insights into their generalization capabilities and adaptability across different scenarios.
	
	\paragraph{Comparison Methods and Evaluation Metrics.} In our quantitative analysis of various dehazing methods, we began by comparing image metrics obtained from training on one dataset and testing on another, to rigorously evaluate each model's adaptability and effectiveness beyond its initial training environment. Our study encompasses a broad spectrum of dehazing techniques, notably the Dark Channel Prior (DCP) \cite{he2010single} method, famed for its reliance on statistical principles and independence from training data. It should be noted that since the Dark Channel Prior method yields results directly without the need for training, we directly applied DCP to the test dataset. The intentional inclusion of DCP as a benchmark was to gauge the performance of our innovative solutions against existing advanced neural network models, such as MixDehazeNet. Utilizing quantitative metrics like Peak Signal-to-Noise Ratio (PSNR) and Structural Similarity Index (SSIM) \cite{wang2004image}, we aimed to assess the success rate of these dehazing techniques, with a particular focus on their generalization capabilities across various scenarios. Additionally, we compared the parameter sizes of different models. Although the size of the model parameter set was considered a secondary factor in our analysis, our primary focus was on elucidating the operational efficiency of each dehazing strategy. Lastly, given dehazing's significance as a crucial preprocessing step for subsequent advanced image tasks, we further tested the effectiveness of different models in actual applications by using the dehazed images for object recognition tasks, thereby evaluating their real-world impact on advanced imaging tasks.
	
	\subsection{Quantitative Analysis of Generalization Across Datasets}
	In our detailed quantitative evaluation, we focus on assessing the generalization capabilities of a broad spectrum of dehazing methods. This involved training models on a specific dataset and then testing them on a distinct one, meticulously examining each model's ability to perform robustly beyond its original training environment. Such an approach enabled us to thoroughly evaluate the adaptability and effectiveness of each method, extending well beyond their initial training parameters. Central to our study was the Dark Channel Prior (DCP) method, celebrated for its reliance on statistical foundations and its unique characteristic of not requiring training data. Through the application of quantitative metrics like Peak Signal-to-Noise Ratio (PSNR) and Structural Similarity Index (SSIM), we aimed to systematically measure the dehazing efficacy of these techniques, with a keen focus on their capacity to adapt across varied conditions. Additionally, we considered the parameter sizes of the models as a secondary yet insightful aspect of our evaluation. Given the pivotal importance of dehazing in preparing images for advanced processing tasks, we further investigated the real-world utility of these models through their application in object recognition tasks, thereby gauging their substantive influence on complex imaging challenges.
	
	\paragraph{Evaluation on Haze4K} In accordance with the experimental setup described earlier, we initially trained our model using the Hazy\_NYU\_DepthV2 dataset and conducted testing on the Haze4K dataset, as outlined in Table \ref{tab:result_Haze4K}. It's noteworthy that PriorNet exhibited remarkable performance in these tests, boasting a PSNR of 17.2756 and an SSIM of 0.8329. These results highlight its exceptional ability to generalize and efficiently tackle diverse dehazing tasks and environmental conditions. In contrast, DCP only achieved modest results with a PSNR of 14.0990 and an SSIM of 0.7588. This disparity underscores the limitations of DCP, which relies heavily on statistical principles and preset haze estimation conditions, leading to reduced performance when real-world conditions deviate from these assumptions. In comparison to the lightweight model AODNet, PriorNet displayed significantly superior performance. While larger models such as DEA Net and MixDehazeNet showed promise in dehazing, their generalization capabilities were comparatively lacking.
	\begin{table}[h!]
		\centering
		\caption{Results on Haze4K.}
		\label{tab:result_Haze4K}
		\begin{tabular}{
				>{\centering\arraybackslash}m{3cm} 
				>{\centering\arraybackslash}m{2cm} 
				S[table-format=2.4, table-number-alignment=center] 
				S[table-format=1.4, table-number-alignment=center] 
			}
			\toprule
			\multirow{2}{*}{\textbf{Method}} & \multirow{2}{*}{\textbf{Venue}} & \multicolumn{2}{c}{\textbf{Metrics}} \\
			\cmidrule(lr){3-4}
			& & {\textbf{PSNR}} & {\textbf{SSIM}} \\
			\midrule
			DCP \cite{he2010single} & TPAMI'10 & 14.0990 & 0.7588 \\
			DEA-Net \cite{chen2024dea} & TIP'24 & 12.9566 & 0.7326 \\
			DehazeFormer \cite{song2023vision} & TIP'23 & 10.8677 & 0.6355 \\
			DFF-Net \cite{wang2024image} & - & 11.8655 & 0.7387 \\
			AOD-Net \cite{li2017aod} & ICCV'17 & 15.1254 & 0.7525 \\
			MixDehazeNet \cite{lu2023mixdehazenet} & IJCNN'24 & 13.0012 & 0.7331 \\
			\textbf {PriorNet (Ours)} & - & \textbf {17.2756} & \textbf {0.8329} \\
			\bottomrule
		\end{tabular}
	\end{table}

	\paragraph{Evaluation on Hazy\_NYU\_DepthV2} To ensure a comprehensive evaluation of the generalization abilities of various dehazing methods, we inverted the roles of the test and training datasets—training on the Haze4k dataset and testing on Hazy\_NYU\_\\DepthV2. The detailed results of these evaluations are meticulously recorded in Table \ref{tab:result_NYU}. Our observations reveal that PriorNet demonstrates remarkable adaptability and performance on unfamiliar datasets, achieving a Peak Signal-to-Noise Ratio (PSNR) of 17.1608 and a Structural Similarity Index (SSIM) of 0.7660. These achievements firmly establish PriorNet as a leading contender among all the neural network-based methods we evaluated. Notably, when compared to AOD-Net, which attained PSNR of 14.5292 and SSIM of 0.7136, PriorNet showcases relatively superior generalization capabilities despite similar parameter counts. However, the performance of other neural network models falls short in comparison. It's noteworthy that the Dark Channel Prior (DCP) method outperforms all models, achieving an impressive PSNR of 18.6280 and SSIM of 0.7819, primarily because the Hazy\_NYU\_DepthV2 dataset adheres to prior conditions.

	\sisetup{
		round-mode          = places, 
		round-precision     = 4, 
		detect-all,         
		table-space-text-post = \textbf{}, 
	}
	\begin{table}[h!]
		\centering
		\caption{Results on Hazy\_NYU\_DepthV2.}
		\label{tab:result_NYU}
		\begin{tabular}{
				>{\centering\arraybackslash}m{3cm} 
				>{\centering\arraybackslash}m{2cm} 
				S[table-format=2.4, table-number-alignment=center] 
				S[table-format=1.4, table-number-alignment=center] 
			}
			\toprule
			\multirow{2}{*}{\textbf{Method}} & \multirow{2}{*}{\textbf{Venue}} & \multicolumn{2}{c}{\textbf{Metrics}} \\
			\cmidrule(lr){3-4}
			& & {\textbf{PSNR}} & {\textbf{SSIM}} \\
			\midrule
			\bfseries DCP & TPAMI'10 & \bfseries 18.6280 & \bfseries 0.7819 \\
			DEA-Net & TIP'24 & 14.8864 & 0.7233 \\
			DehazeFormer & TIP'23 & 9.3744 & 0.5908 \\
			DFF-Net & - & 11.4134 & 0.7302 \\
			AOD-Net & ICCV'17 & 14.5292 & 0.7136 \\
			MixDehazeNet & IJCNN'24 & 13.9572 & 0.7168 \\
			PriorNet (Ours) & - &  17.1680 &  0.7660 \\
			\bottomrule
		\end{tabular}
	\end{table}
	
	\subsection{Comparative Analysis of Model Parameter Sizes}
	
	The data presented in Table \ref{tab:model_parameter_sizes} unequivocally illustrates the substantial advantage of PriorNet in terms of the number of model parameters. Relative to extant technologies, the parameter count of most models is at least two orders of magnitude greater than that of our proposed method. This disparity is even more pronounced in the realm of multi-task processing, where the current state-of-the-art model, MixDehazeNet, possesses a parameter volume that surpasses ours by three orders of magnitude. This marked reduction in parameter count not only signifies considerable savings in computational resources and storage capacity but also facilitates the deployment on edge devices. It is particularly noteworthy that despite the drastic reduction in parameter numbers, PriorNet demonstrably outperforms all current models on the market in terms of generalization performance, a fact that our experimental results have robustly validated.
	
	\begin{table}[h!]
		\centering
		\caption{Model Parameter Sizes of Dehazing Methods}
		\label{tab:model_parameter_sizes}
		\begin{tabular}{lc}
			\toprule
			Methods & \multicolumn{1}{l}{Size (Kb)} \\ 
			\midrule
			MixDehazeNet & 143387 \\
			DEA-Net & 14331 \\
			DFF-Net & 68739 \\
			DehazeFormer & 5593 \\
			AOD-Net & 11 \\
			PriorNet (Ours) & 18 \\
			\bottomrule
		\end{tabular}
	\end{table}
	
	\subsection{Ablation Studies}
	
	In the ablation studies we conducted, by sequentially removing key components from PriorNet, we explored their specific impact on the performance of image dehazing. We focused on evaluating the effects of removing the Multidimensional Interactive Attention (MIA) mechanism and employing non-uniform convolutional kernel sizes. The results, as shown in Table \ref{tab:ablation}, conclusively highlight the central role of the MIA mechanism in enhancing the model's dehazing effectiveness; concurrently, the use of uniform convolutional kernel sizes and a simplified network architecture collectively contribute to the model's efficiency and generalization capabilities. This series of experiments not only confirms the importance of each element within the design of PriorNet but also emphasizes their interplay, providing a solid foundation for efficiently and effectively addressing the problem of image dehazing.
	
	\begin{table}[h!]
		\centering
		\caption{Ablation Study Results on PriorNet}
		\label{tab:ablation}
		\begin{tabular}{
				>{\centering\arraybackslash}m{3cm} 
				S[table-format=2.4] 
				S[table-format=1.4] 
			}
			\toprule
			\textbf{Method} & {\textbf{PSNR}} & {\textbf{SSIM}} \\
			\midrule
			Only 3 * 3 & 13.3155 & 0.6897  \\
			Only 5 * 5 & 14.6674  & 0.7156  \\
			CA + 5 * 5 & 15.2460  & 0.7201 \\
			MIA + Multi\_coresize & 14.5292 & 0.7136 \\
			\bottomrule
		\end{tabular}
	\end{table}

	\begin{figure*}
		\centering
		\includegraphics[width=\textwidth]{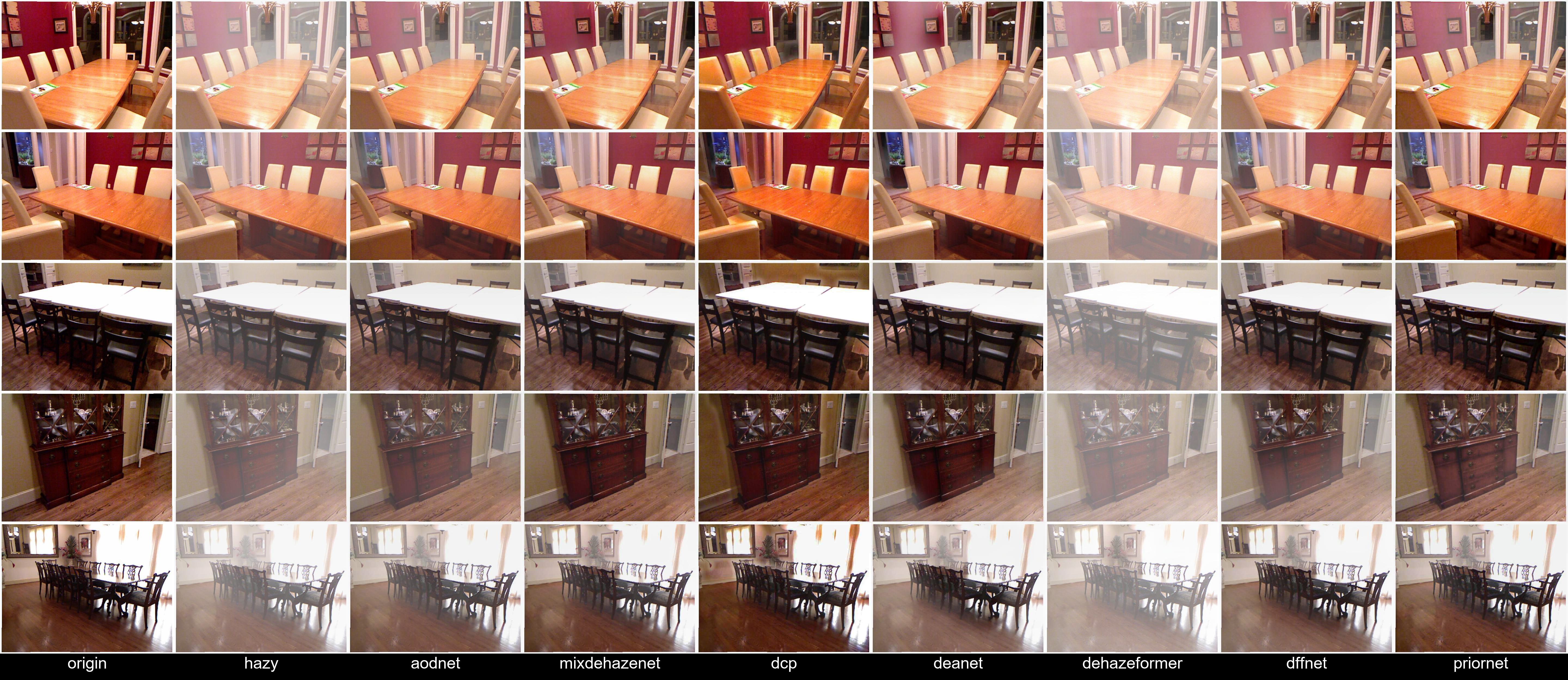}
		\caption{Visual Results of Dehazing Methods.}
		\label{fig:final_result}
	\end{figure*}
	
	\subsection{The Influence of Image Dehazing on Future Vision Tasks}
	To comprehensively evaluate the quality of images processed by various dehazing methods, we employed a YOLOv4 \cite{bochkovskiy2020yolov4} model trained on the COCO dataset \cite{lin2014microsoft} to identify objects within the processed images from the Hazy\_NYU\_DepthV2 dataset. As shown in the table \ref{tab:Accuracy}, PriorNet clearly demonstrates the best performance, closely followed by DCP. However, it's worth noting that as the image quality improves, the enhancement in object recognition tasks from each method appears to encounter a bottleneck. 
	\begin{table}[h!]
		\centering
		\caption{Ablation Study Results on PriorNet}
		\label{tab:Accuracy}
		\begin{tabular}{
				>{\centering\arraybackslash}m{3cm} 
				S[table-format=2.4] 
				S[table-format=1.4] 
			}
			\toprule
			\textbf{Method} & {\textbf{Accuracy}}\\
			\midrule
			DCP & 0.6612 \\
			MixDehazeNet & 0.5156 \\
			DEA-Net & 0.6122 \\
			DFF-Net & 0.6166 \\
			DehazeFormer & 0.6134 \\
			AOD-Net & 0.6565 \\
			\bfseries PriorNet (Ours) & \bfseries 0.6732 \\
			\bottomrule
		\end{tabular}
	\end{table}
	\section{Visual Results and Analysis}
	Figure \ref{fig:final_result} presents a comparative analysis of the efficacy of various dehazing techniques, clearly illustrating that aside from DeHazeFormer, the performance of other methodologies on the Hazy\_NYU\\\_DepthV2 dataset is quite notable. However, a critical observation reveals that DCP encounters severe color distortion due to its reliance on built-in prior assumptions, leading to pronounced halos and light patches near object edges, significantly impeding the accuracy and implementation complexity of subsequent tasks. Moreover, while DFFNet and AOD-Net have made strides in reducing haze, they consistently generate images with diminished brightness, indicating that the haze effects are not fully eradicated. DFFNet shows overall better results compared to AOD-Net, yet some areas with low luminance still suffer from inadequate dehazing. On the other hand, DEA-Net and MixDehazeNet struggle to maintain clarity at the edges or textures of objects, resulting in noticeable blurring that compromises the sharpness and detail of the images.In contrast, Priornet distinguishes itself by offering the most outstanding overall visual experience in dehazing. Our network significantly showcases its superior performance in processing image details and maintaining color fidelity, illustrating its exceptional capabilities in enhancing visual clarity and vibrancy.

	\section{Conclusion}
	In this study, we draw upon the principles of mathematical models and merge them with the potent abstraction capabilities of neural networks to conceive an innovative attention mechanism. This mechanism integrates channel and spatial attention strategies, aimed at effectively capturing the coarse features present in complex real-world scenarios, and is designated as the Multidimensional Interactive Attention (MIA). Building upon MIA, we further introduce Priornet—a lightweight dehazing network characterized by its remarkable generalization ability, demonstrating superior performance in the task of single-image dehazing.The architecture of Priornet is both simple and efficient, ensuring high accuracy in dehazing operations while its lightweight nature facilitates effortless deployment on edge devices. Extensive testing across various datasets has validated Priornet's formidable capability in haze removal and clarity restoration, alongside its exceptional preservation of image details and color fidelity. Overall, our work not only advances the field of image dehazing but also offers a new perspective and tool for related domains, especially in the pursuit of generalization and deployability.

	\bibliographystyle{ACM-Reference-Format}
	\bibliography{sample-base}


\end{document}